# A transformer-based approach to video frame-level prediction in Affective Behaviour Analysis In-the-wild

Dang-Khanh Nguyen
Chonnam National University
Gwangju, South Korea
khanhnd185@gmail.com

Ngoc-Huynh Ho
Chonnam National University
Gwangju, South Korea
nhho@chonnam.ac.kr

Sudarshan Pant
Chonnam National University
Gwangju, South Korea
sudarshan@chonnam.ac.kr

Hyung-Jeong Yang*
Chonnam National University
Gwangju, South Korea
hjyang@jnu.ac.kr

## Abstract

*In recent years, transformer architecture has been a dominating paradigm in many applications, including affective computing. In this report, we propose our transformer-based model to handle Emotion Classification Task in the 5th Affective Behavior Analysis In-the-wild Competition. By leveraging the attentive model and the synthetic dataset, we attain a score of 0.4775 on the validation set of Aff-Wild2, the dataset provided by the organizer.*

## 1. Introduction

Representing human emotion is a fundamental topic in behavior analysis [3,6]. A naïve approach is using a discrete classification of 6 or 7 basic emotions. Besides, emotion could be also represented in a continuous 2-dimensional space (i.e., valence-arousal). Other proposals use the presentation of the facial action units as the emotional representation. Analyzing human interaction from various perspectives [11,14] can help researchers deeply understand their feeling and behavior.

The 5th Competition on Affective Behavior Analysis In-the-wild (ABAW5) aims to achieve that target by conducting three challenges corresponding to three methods representing the human emotion: Valence-Arousal (VA) estimation [9,10], Expression (EX) recognition, and Action Unit (AU) detection [8]. Hume AI also collaborates with the organizer on the Emotional Reaction Intensity (ERI) estimation challenge, which is a brand-new task compared to the three above traditional ones [2,4,5]. This is also a regression problem measuring the level of each emotion label.

This paper introduces our solution for the three conventional challenges of the competition. We adopt a pre-trained EfficientNet [15] as a feature extractor and manipulate a transformer encoder [12] to learn the relationship among a sequence of facial frames. Noticeably, in Expression classification task, to resolve the imbalance limitation, we utilize the generated data from ABAW4's Learning from Synthetic Data (LSD) challenge [1] to increase the scale of the training dataset. It considerably improves the performance of our discriminative model.

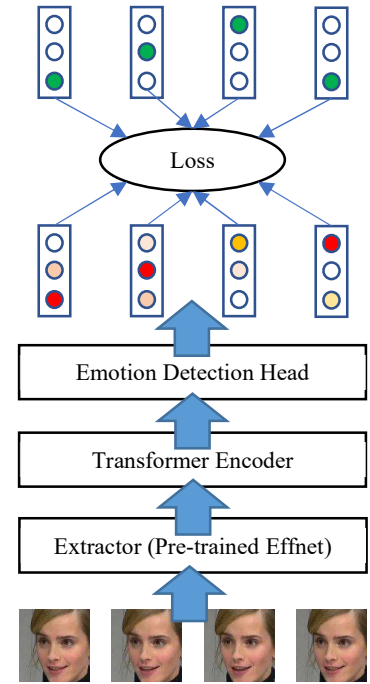

Figure 1: The block diagram of our proposed method. The vectors in green are the one-hot vectors of the ground truth, and the vectors in red are the logit prediction of the models. Each vector corresponds to an image in the sequence of frames.

## 2. Proposed Method

We utilize the transformer paradigm for our proposed method. To create a sequence of embeddings from a series of facial frames, we exploit the pre-trained EfficientNet [13] well trained with multiple facial analysis tasks. This pre-trained network extracts a feature vector for each image. Then, the features are projected to the embedding space by a linear layer before being fed to a transformer encoder. The encoder is designed similarly to the conventional transformer encoder [12] with multi-head attention and feed-forward networks. The output from the encoder is sent to the emotion detection head, which is simply a fully-connected neural network with one hidden layer.

In the classification task, to overcome the imbalance issue of the dataset, we apply weighted scores for each emotion in the cross-entropy loss. On the other hand, due to the quadratic complexity of the input length of the attention mechanism, we cannot put all frames of a video into the transformer. The frame sequence is fragmented into multiple segments with a common length. The detail of our proposed method is illustrated in Figure 1.

## 3. Dataset and Experiment

### 3.1. Dataset

We used Aff-Wild2 [7] and the synthetic facial dataset [1] for our training process. The Aff-Wild2 for expression classification is a collection of 546 videos. More than 2 million frames are extracted from the videos and annotated as one among 8 emotional classes. The classification comprises 6 basic emotions: "anger", "disgust", "fear", "happiness", "sadness" and "surprise". Two additional classes are included, "neural" and "other". They are also the two classes having the largest number of samples in the dataset.

To enlarge the scale of the training dataset, we leverage the synthetic facial dataset from the 4[th] ABAW competition, Learning from Synthetic Data challenge. The corpus includes 277,251 synthesized images classified into 6 basic classes. We merge this dataset with the training split of the Aff-Wild2 dataset. It does not only increase the size of training samples but also decreases the influence of the imbalance issue.

The Aff-Wild2 is provided in three versions: raw videos, cropped images, and cropped-aligned images. We used the cropped-aligned images in our experiments. These images have the same size of 112x112 pixels. Regarding the synthetic data, the image size is 128x128 pixels.

### 3.2. Experiment setting

The sequence of frames in each video is split into segments comprising 64 frames. All images are normalized and resized to 224x224 pixels. Regarding the transformer, we use a hidden size and feed-forward size of 512, dropout ratio is 0.1. The emotion detection head is a neural network with one hidden layer with a size of 256. We train our models in 20 epochs with a batch size of 64 and a learning rate of 0.001. The Adam optimizer is used with the weight decay of $\frac{1}{64}$.

To boost the performance, we apply the ensemble of 3 models with different configurations of the transformer encoder. The number of heads and layers in 3 settings are (4, 4), (8, 4), and (4, 6), respectively. Soft average voting is used to obtain the final prediction from 3 logit vectors of the models.

## 4. Results

We compare our models in various settings. Particularly, the performance on the validation set of each approach in Expression recognition challenge is shown in table 1. As the result, using a pre-trained EfficientNet followed by fully connected layers obtains a score of 0.3327, which is higher than the baseline [16]. Adding a transformer encoder to the model can improve the F1-score. Moreover, if we exploit the synthetic data to train the transformer encoder, the results are boosted significantly to more than 0.44. Then, we try to increase the scale of the transformer, particularly, the number of heads and encoder layers. The performance is slightly enhanced but not noticeable.

**Table 1** The performance on the validation set of various settings. Effnet and FCs stand for Pre-trained EfficientNet and Fully Connected Layers. N and h are number of layers and heads in transformer encoder. "Syn" implies that the experiments use synthetic data in the training process.

| Configuration | F1-score |
|---|---|
| Effnet+FCs | 0.3327 |
| Effnet+ Encoder (N=4, h=4) +FCs | 0.3615 |
| (1) Effnet+ Encoder (N=4, h=4)+FCs+Syn | 0.4400 |
| (2) Effnet+Encoder (N=4, h=8)+FCs+Syn | 0.4424 |
| (3) ffnet+Encoder (N=6, h=4)+FCs+Syn | **0.4555** |

Afterward, we use an ensemble to combine the logit output of each model settings. Consequently, we attain better output compared to a single model. The combinations of two transformer-based models attain the F1-score from 0.4663 to 0.4729. The ensemble of 3 best configurations accomplishes a score of 0.4775. This is also our best result on the validation set of Aff-Wild2.

**Table 2** The results of ensembles of models on validation set. (1), (2), (3) denote the configurations mentioned in Table 1.

| Ensemble model | F1-score |
|---|---|
| Soft average voting (1) (2) | 0.4663 |
| Soft average voting (1) (3) | 0.4672 |
| Soft average voting (3) (2) | 0.4729 |
| Soft average voting (1), (2) and (3) | **0.4775** |

Regarding the two remained challenges, we apply the same methodology and configuration except that the synthetic dataset is not used. Again, the metric scores are enhanced after we fuse the prediction of multiple models by soft average voting. The detailed results of AU detection and VA estimation tasks are denoted in Table 3 and Table 4, respectively.

**Table 3** The results of our proposed method on the validation set of Aff-Wild2 in AU detection task.

| Ensemble model | F1-score |
| --- | --- |
| (1) Effnet+ Encoder (N=4, h=4)+FCs | 0.51696 |
| (2) Effnet+Encoder (N=4, h=8)+FCs | 0.51146 |
| (3) Effnet+Encoder (N=6, h=4)+FCs | 0.51192 |
| Soft average voting (1) (2) | 0.51960 |
| Soft average voting (1) (3) | 0.52021 |
| Soft average voting (3) (2) | 0.51709 |
| Soft average voting (1), (2) and (3) | **0.52085** |

**Table 3** The results of our proposed method on the validation set of Aff-Wild2 in VA estimation task.

| Ensemble model | F1-score |
| --- | --- |
| Effnet+ Encoder (N=4, h=4)+FCs (1) | 0.48296 |
| Effnet+Encoder (N=4, h=8)+FCs (2) | 0.48819 |
| Effnet+Encoder (N=6, h=4)+FCs (3) | 0.47389 |
| Soft average voting (1) (2) | 0.49684 |
| Soft average voting (1) (3) | 0.49679 |
| Soft average voting (3) (2) | 0.49874 |
| Soft average voting (1), (2) and (3) | **0.50290** |

## 5. Conclusion

In this report, we provide a straightforward and effective method for frame-level video classification. The transformer is employed to learn the correlation among the frames in a sequence. In the emotion classification task of the ABAW5 competition, our best achievement on the validation set of Aff-Wild2 is 0.4775, which outperforms the baseline provided by the organizer.

## References

[1] Kollias, Dimitrios. ABAW: Learning from Synthetic Data & Multi-Task Learning Challenges. arXiv preprint arXiv:2207.01138 (2022).
[2] Kollias, Dimitrios. Abaw: Valence-arousal estimation, expression recognition, action unit detection & multi-task learning challenges. In Proceedings of the IEEE/CVF Conference on Computer Vision and Pattern Recognition (pp. 2328–2336), 2022.
[3] Kollias, Dimitrios, Viktoriia, Sharmanska, and Stefanos, Zafeiriou. Distribution Matching for Heterogeneous Multi-Task Learning: a Large-scale Face Study. arXiv preprint arXiv:2105.03790 (2021).
[4] Kollias, Dimitrios, and Stefanos, Zafeiriou. Analysing affective behavior in the second abaw2 competition. In Proceedings of the IEEE/CVF International Conference on Computer Vision (pp. 3652–3660), 2021.
[5] Kollias, Dimitrios, and Stefanos, Zafeiriou. Affect Analysis in-the-wild: Valence-Arousal, Expressions, Action Units and a Unified Framework. arXiv preprint arXiv:2103.15792 (2021).
[6] Kollias, D, A, Schulc, E, Hajiyev, and S, Zafeiriou. Analysing Affective Behavior in the First ABAW 2020 Competition. In 2020 15th IEEE International Conference on Automatic Face and Gesture Recognition (FG 2020)(FG) (pp. 794–800), 2020.
[7] Kollias, Dimitrios, and Stefanos, Zafeiriou. Expression, Affect, Action Unit Recognition: Aff-Wild2, Multi-Task Learning and ArcFace. arXiv preprint arXiv:1910.04855 (2019).
[8] Kollias, Dimitrios, Viktoriia, Sharmanska, and Stefanos, Zafeiriou. Face Behavior a la carte: Expressions, Affect and Action Units in a Single Network. arXiv preprint arXiv:1910.11111 (2019).
[9] Kollias, Dimitrios, Panagiotis, Tzirakis, Mihalis A, Nicolaou, Athanasios, Papaioannou, Guoying, Zhao, Bjorn, Schuller, Irene, Kotsia, and Stefanos, Zafeiriou. Deep affect prediction in-the-wild: Aff-wild database and challenge, deep architectures, and beyond. International Journal of Computer Vision (2019): 1–23.
[10] Zafeiriou, Stefanos, Dimitrios, Kollias, Mihalis A, Nicolaou, Athanasios, Papaioannou, Guoying, Zhao, and Irene, Kotsia. Aff-wild: Valence and arousal 'in-the-wild' challenge. In Computer Vision and Pattern Recognition Workshops (CVPRW), 2017 IEEE Conference on (pp. 1980–1987), 2017.
[11] Nguyen, Dang-Khanh, Sudarshan Pant, Ngoc-Huynh Ho, Guee-Sang Lee, Soo-Hyung Kim, and Hyung-Jeong Yang. Affective Behavior Analysis Using Action Unit Relation Graph and Multi-task Cross Attention. In European Conference on Computer Vision, pp. 132-142. Springer, Cham, 2023.
[12] Vaswani, Ashish, Noam Shazeer, Niki Parmar, Jakob Uszkoreit, Llion Jones, Aidan N. Gomez, Łukasz Kaiser, and Illia Polosukhin. Attention is all you need. Advances in neural information processing systems 30 (2017).
[13] Savchenko, Andrey V. Frame-level prediction of facial expressions, valence, arousal and action units for mobile devices. arXiv preprint arXiv:2203.13436 (2022).
[14] Zhang, Tenggan, Chuanhe Liu, Xiaolong Liu, Yuchen Liu, Liyu Meng, Lei Sun, Wenqiang Jiang, Fengyuan Zhang, Jinming Zhao, and Qin Jin. Multi-Task Learning Framework for Emotion Recognition In-the-Wild. In European Conference on Computer Vision, pp. 143-156. Springer, Cham, 2023.
[15] Tan, Mingxing, and Quoc Le. Efficientnet: Rethinking model scaling for convolutional neural networks. In International conference on machine learning, pp. 6105-6114. PMLR, 2019.
[16] Kollias, Dimitrios, Panagiotis, Tzirakis, Alice, Baird, Alan, Cowen, and Stefanos, Zafeiriou. ABAW: Valence-Arousal Estimation, Expression Recognition, Action Unit Detection & Emotional Reaction Intensity Estimation Challenges. arXiv preprint arXiv:2303.01498 (2023).